\documentclass[11pt]{article}

\usepackage[final]{acl}

\usepackage{fancyhdr}
\fancypagestyle{arxivnote}{%
  \fancyhf{}%
  \fancyhead[C]{\small\itshape Accepted at EMNLP 2026}%
}
\usepackage{times}
\usepackage{latexsym}

\usepackage[T1]{fontenc}

\usepackage[utf8]{inputenc}

\usepackage{microtype}

\usepackage{inconsolata}

\usepackage{graphicx}

\usepackage{amsmath,amssymb}
\usepackage{booktabs}
\usepackage{multirow}
\usepackage{enumitem}
\usepackage{xcolor}
\usepackage{float}
\usepackage{tabularx}
\newcolumntype{L}{>{\raggedright\arraybackslash}X}

\title{Better Together: Complementary Query Rewriting 
Under a Strong RAG Baseline}

\author{Sara Shanian \quad Xiaoqin Yi \quad Pavlo Ruban \quad Kurt MacDonald \\
  ServiceNow}

\begin{document}
\maketitle
\thispagestyle{arxivnote}

\begin{abstract}
A popular way to improve Retrieval-Augmented Generation (RAG) is to rewrite the user's question into several variants and search with all of them. We test whether this actually helps once the underlying search is already strong. Under one fixed, competitive pipeline (BGE dense retrieval, cross-encoder reranking, and MMR diversification), we compare four query-rewriting strategies (S1--S4) against two strong LLM baselines (HyDE, Query2Doc) on three datasets (HotpotQA, AmbigNQ, and the 512K-document EnterpriseRAG-Bench) over three seeds with paired-bootstrap significance tests.

Our headline result is that \textbf{rewriting alone is at best competitive with a strong baseline, but combining methods yields outsized gains because different strategies fail on different questions.} A post-hoc union of four methods (S1+S3+S4+HyDE; seed-42 analysis) improves HIT@10 over the dense baseline by +12.5 points on enterprise-style data (51.70 vs 39.22), and a five-method union reaches 52.98 (+13.8 over baseline). Budget-matched controls (TopK-100 reranking, multi-pass HNSW) capture only $\sim$40\% of this gain, confirming that complementarity, not retrieval budget, is the primary driver. On HotpotQA the union adds +1.6 to +1.8 points ($p < 0.001$), saturating the all-method oracle; on AmbigNQ the same fusion \emph{hurts} ($-2.4$ below the best solo, $p < 0.001$), and we analyze when and why. Because rewriting is expensive (one LLM call each), we turn this into a cost-aware heuristic evaluated in simulation: a confidence-gated router that runs rewriting only when the baseline's own top-1 score is low. It captures about half of the enterprise full-merge gain (+4.3 HIT@10 over the baseline) while paying the rewriting cost on $<$40\% of queries, and automatically declines to rewrite on AmbigNQ, sidestepping the loss that blind merging causes. A downstream answer-quality evaluation confirms that the router improves F1 by +1.92 ($p < 0.01$) at roughly 40\% of the expansion cost. \textbf{In short: treat query rewriting as a complementary coverage source, applied through cost-aware routing, not as a standalone replacement for a strong baseline.}
\end{abstract}

\section{Introduction}

Retrieval-Augmented Generation (RAG) grounds Large Language Model (LLM) answers in documents retrieved from a corpus \citep{lewis2020retrieval}. A widely held belief, baked into popular tooling \citep{langchain2025multiqueryretriever}, is that searching with a \emph{single} query is brittle, and that \emph{rewriting} the query into several variants improves coverage \citep{li2024dmqrrag,zhang2025levelrag}.

\noindent We put this belief to the test, and add one missing ingredient: \textbf{a strong baseline}. Most prior comparisons rewrite on top of a weak retriever (a bare bi-encoder, no reranking), which makes almost any rewrite look good. We instead fix a competitive modern pipeline: BGE dense retrieval, a cross-encoder reranker, and Maximal Marginal Relevance (MMR) diversification. We then ask: \emph{does rewriting still help on top of that?}

\noindent In plain terms, the answer is:

\begin{enumerate}[noitemsep,leftmargin=*,label=(\roman*)]
\item \textbf{Alone, rewriting is competitive at best.} On all three datasets, multi-query rewriting strategies (S2--S4) match or modestly beat the no-LLM baseline (S1), and LLM \emph{document}-generation methods (HyDE, Query2Doc) perform similarly. No single strategy dominates across datasets.
\item \textbf{Together, they win.} Methods fail on \emph{different} questions, so merging them beats the best single method, hugely so on enterprise data: a post-hoc ensemble reaches 52.98\% HIT@10 (+13.8 over the S1 baseline), within $\sim$2.8 points of the all-method oracle (55.74). Budget-matched baselines capture only $\sim$40\% of this gain, confirming that complementarity, not retrieval depth, is the driver.
\end{enumerate}

\paragraph{Contributions.}
(1) A controlled, statistically rigorous comparison of six methods under one fixed strong pipeline, three datasets, three seeds (Sections~\ref{sec:strategies}--\ref{sec:solo}).
(2) The complementarity result: rewriting helps by \emph{combination}, not \emph{replacement}, validated against budget-matched controls (Sections~\ref{sec:complementarity}--\ref{sec:ensemble}).
(3) A negative result with its mechanism: why combining fails on AmbigNQ (Section~\ref{sec:discussion}).
(4) A cost-aware, confidence-gated router that, in simulation, realizes about half of the enterprise full-merge gain (+4.3 HIT@10) at $<$40\% of the rewriting cost and self-disables where rewriting hurts (Sections~\ref{sec:cost}--\ref{sec:router}).
(5) A downstream answer-quality evaluation showing that the router improves answer F1 by +1.92 ($p < 0.01$) over the baseline while preserving near-full-merge quality.

\section{Related Work}

\paragraph{RAG.} RAG \citep{lewis2020retrieval} pairs LM knowledge with retrieval. Dense Passage Retrieval \citep{karpukhin2020dense}, Fusion-in-Decoder \citep{izacard2021leveraging}, and Self-RAG \citep{asai2023self} advanced retrieval and evidence use. Systematic studies show RAG quality hinges on pipeline choices like reranking and chunking \citep{zhang2024searching}, which is exactly why we hold the pipeline \emph{fixed} and vary only the rewriting strategy.

\paragraph{Query rewriting.} Classic relevance feedback expands queries with retrieved terms \citep{rocchio1971relevance}; neural methods generate expansions \citep{nogueira2019document}; decomposition aids multi-hop reasoning \citep{perez2020unsupervised}. Two LLM \emph{document}-generation baselines are now standard: HyDE retrieves with a generated hypothetical answer \citep{gao2022precise}, and Query2Doc concatenates the query with a generated pseudo-document \citep{wang2023query2doc}. Recent RAG work produces diverse rewrites \citep{li2024dmqrrag}, plans multi-hop logic \citep{zhang2025levelrag}, or optimizes rewrite choice under budget \citep{yang2025mdp_mqr_rag}. We differ by measuring these methods \emph{against a strong baseline} and analyzing their \emph{complementarity} rather than their solo accuracy. Reranking \citep{nogueira2019passage} and MMR \citep{carbonell1998use} are part of our fixed pipeline.

\section{Reformulation Strategies}
\label{sec:strategies}

We study six methods in three families: a \textbf{no-LLM baseline} (S1); \textbf{multi-query rewriting} (S2--S4, which generate alternative \emph{queries}); and \textbf{document generation} (HyDE, Query2Doc, which generate a pseudo-\emph{document}). The key practical axis is \textbf{cost}: S1 makes no LLM call, while every other method makes one LLM call per query (S2/S3 also issue several retrievals). Table~\ref{tab:strategies} summarizes them, using the running query \emph{``What are the side effects of Tylenol?''}.

\begin{table*}[t]
\centering
\small
\begin{tabularx}{\textwidth}{@{}l L L L@{}}
\toprule
\textbf{Strategy} & \textbf{What it does} & \textbf{Cost} & \textbf{Example for ``side effects of Tylenol?''} \\
\midrule
S1 Parent      & Use the original query, unchanged        & \textbf{Cheap (no LLM)}      & \emph{(query used as-is)} \\
S2 Neighbor    & Ask sibling questions (related entities)  & Expensive (LLM + $n$ retr.)   & ``side effects of ibuprofen?'' \\
S3 Synonym     & Reword with different vocabulary          & Expensive (LLM; thesaurus-able)& ``adverse reactions to acetaminophen?'' \\
S4 Comparative & Compare the subject vs.\ alternatives     & Expensive (LLM, 1 variant)    & ``Tylenol vs.\ Advil side effects?'' \\
HyDE           & Write a fake answer passage, search with it & Expensive (LLM)             & \emph{(pseudo-passage on acetaminophen)} \\
Query2Doc      & Add a generated pseudo-document to the query & Expensive (LLM)             & \emph{(query $+$ pseudo-passage)} \\
\bottomrule
\end{tabularx}
\caption{The six methods. Only S1 avoids an LLM call. Among rewriting strategies, S4 is cheapest (one variant); S3's intent (synonyms) is the most amenable to an LLM-free (dictionary/WordNet) implementation, though we use the LLM version for fairness.}
\label{tab:strategies}
\end{table*}

\paragraph{S1: Parent (baseline).} Search with the original query unchanged. \textbf{Cheap:} no LLM call. This is the strong reranked baseline all others are measured against.

\paragraph{S2: Neighbor.} The LLM proposes related ``sibling'' questions (e.g., \emph{``side effects of ibuprofen?''}), each retrieving documents that are merged. \textbf{Expensive}, and the highest-variance strategy; it generates the most query variations.

\paragraph{S3: Synonym.} The LLM restates the question with different words (e.g., \emph{``adverse reactions to acetaminophen?''}), targeting vocabulary mismatch. \textbf{Expensive}, but the one strategy a thesaurus could approximate without an LLM.

\paragraph{S4: Comparative.} The LLM forms a comparison question (e.g., \emph{``how do Tylenol's side effects compare to Advil's?''}). \textbf{Expensive, but cheapest of S2--S4} (one variant), and the most useful for combining.

\paragraph{HyDE / Query2Doc.} HyDE \citep{gao2022precise} writes a hypothetical answer passage and searches with its embedding; Query2Doc \citep{wang2023query2doc} appends a generated pseudo-document to the query. \textbf{Expensive}, and strong solo methods.

\paragraph{Merging.} For multi-query strategies, variants are quality-filtered (cosine to the original in $[0.4,0.9]$), each retrieves documents, and results are merged by \emph{score-based mixing} (keep each document's max similarity across variants, take the global top-$K$). The same merge forms cross-strategy \emph{ensembles} in Section~\ref{sec:ensemble}.

\section{Experimental Setup}
\label{sec:setup}

\paragraph{Datasets} (Table~\ref{tab:datasets}). \textbf{HotpotQA} \citep{yang2018hotpotqa}: multi-hop questions; we use the first 5{,}000 examples of the training split in file order (no shuffle). \textbf{AmbigNQ} \citep{min2020ambigqa}: ambiguous questions with multiple interpretations; the full development set (2{,}002 queries). \textbf{EnterpriseRAG-Bench} \citep{onyx2024enterpriseragbench,enterpriserag2026arxiv}: $\sim$512K synthetic enterprise documents with 470 questions, released May 2026 under MIT license (Hugging Face: \texttt{onyx-dot-app/EnterpriseRAG-Bench}); we evaluate on the 470 answerable questions (the benchmark's 30 unanswerable questions have no relevance judgments and are excluded); it is by far the hardest corpus ($\sim$60\% of queries have no relevant document in the baseline's top-10). The benchmark uses ``Redwood Inference,'' a simulated company, as its enterprise scenario. Both public web-QA benchmarks predate the retriever and LLM training cutoffs (HotpotQA 2018, AmbigNQ 2020), so memorization may inflate absolute scores on these datasets; our enterprise benchmark is not subject to this concern.

\begin{table}[!htbp]
\centering
\small
\begin{tabular}{@{}llll@{}}
\toprule
\textbf{Dataset} & \textbf{Docs} & \textbf{Queries} & \textbf{Character} \\
\midrule
HotpotQA            & 49,708  & 5,000 & Multi-hop \\
AmbigNQ             & 28,378  & 2,002 & Ambiguous \\
EnterpriseRAG-Bench & 511,958 & 470   & Enterprise \\
\bottomrule
\end{tabular}
\caption{Evaluation datasets. HotpotQA uses the first 5{,}000 train examples (file order); AmbigNQ is the full dev set; Enterprise uses all 470 queries.}
\label{tab:datasets}
\end{table}

\paragraph{Fixed pipeline.} Every configuration shares: \texttt{bge-base-en-v1.5} embeddings; \texttt{bge-reranker-base}; MMR ($\lambda{=}0.85$); sentence chunking (512/50); score-based mixing; GPT-4.1 (temp.\ 0.7) for all LLM generation. This is a deliberately \emph{strong} baseline, since prior rewriting studies often omit reranking and MMR; cross-encoder reranking is known to substantially close bi-encoder quality gaps~\citep{nogueira2019passage}. Full settings: Appendix~\ref{sec:appendix-config}.

\paragraph{Protocol.} Three seeds (42, 123, 2024); we report mean$\pm$std. We use standard IR metrics and focus on Recall@10, NDCG@10, MRR@10, HIT@10. Significance vs.\ S1 uses a paired bootstrap (2{,}000 iterations, two-sided): $^{*}p{<}0.05$, $^{**}p{<}0.01$, $^{***}p{<}0.001$. Solo results are three-seed means; ensemble and oracle numbers are post-hoc seed-42 analyses (score-based union of each method's top-20 pool, taking top-10).

\section{Results}
\label{sec:results}

\subsection{Alone, Rewriting Is Competitive at Best}
\label{sec:solo}

Table~\ref{tab:main} is the centerpiece: each method's headline metrics per dataset, with significance vs.\ S1.

\begin{table*}[t]
\centering
\small
\begin{tabular*}{\textwidth}{@{\extracolsep{\fill}}llllll@{}}
\toprule
\textbf{Dataset} & \textbf{Strategy} & \textbf{Recall@10} & \textbf{NDCG@10} & \textbf{MRR@10} & \textbf{HIT@10} \\
\midrule
\multirow{6}{*}{HotpotQA}
& S1 (Parent)      & 87.70\,$\pm$0.08 & 84.21\,$\pm$0.08 & 91.49\,$\pm$0.09 & 97.13\,$\pm$0.11 \\
& S2 (Neighbor)    & 86.34\,$\pm$0.10$^{***}$ & 80.51\,$\pm$0.08$^{***}$ & 88.93\,$\pm$0.09$^{***}$ & \textbf{97.65}\,$\pm$0.01$^{**}$ \\
& S3 (Synonym)     & 86.14\,$\pm$0.08$^{***}$ & 80.29\,$\pm$0.10$^{***}$ & 88.62\,$\pm$0.16$^{***}$ & 97.40\,$\pm$0.18 \\
& S4 (Comparative) & 86.27\,$\pm$0.11$^{***}$ & 80.50\,$\pm$0.10$^{***}$ & 88.80\,$\pm$0.11$^{***}$ & 97.33\,$\pm$0.09 \\
& HyDE             & 88.97\,$\pm$0.09$^{***}$ & 85.06\,$\pm$0.06$^{***}$ & 91.62\,$\pm$0.08 & 97.19\,$\pm$0.05 \\
& Query2Doc        & \textbf{89.74}\,$\pm$0.04$^{***}$ & \textbf{85.57}\,$\pm$0.08$^{***}$ & \textbf{91.65}\,$\pm$0.20 & 97.31\,$\pm$0.22 \\
\midrule
\multirow{6}{*}{AmbigNQ}
& S1 (Parent)      & 79.48\,$\pm$0.03 & 59.27\,$\pm$0.02 & 60.58\,$\pm$0.02 & 93.67\,$\pm$0.03 \\
& S2 (Neighbor)    & \textbf{81.71}\,$\pm$0.02$^{***}$ & \textbf{67.29}\,$\pm$0.02$^{***}$ & 71.26\,$\pm$0.02$^{***}$ & 95.44\,$\pm$0.03$^{***}$ \\
& S3 (Synonym)     & 81.65\,$\pm$0.04$^{***}$ & 67.27\,$\pm$0.02$^{***}$ & \textbf{71.28}\,$\pm$0.01$^{***}$ & \textbf{95.47}\,$\pm$0.03$^{***}$ \\
& S4 (Comparative) & 81.59\,$\pm$0.04$^{***}$ & 67.19\,$\pm$0.04$^{***}$ & 71.17\,$\pm$0.06$^{***}$ & 95.34\,$\pm$0.06$^{**}$ \\
& HyDE             & 80.94\,$\pm$0.13$^{***}$ & 60.23\,$\pm$0.03$^{***}$ & 61.16\,$\pm$0.09$^{**}$ & 94.91\,$\pm$0.17$^{***}$ \\
& Query2Doc        & 80.58\,$\pm$0.20$^{**}$ & 60.03\,$\pm$0.10$^{***}$ & 61.08\,$\pm$0.09$^{**}$ & 94.69\,$\pm$0.15$^{**}$ \\
\midrule
\multirow{6}{*}{Enterprise}
& S1 (Parent)      & 34.33\,$\pm$0.39 & 31.99\,$\pm$0.36 & 34.05\,$\pm$0.39 & 39.22\,$\pm$0.49 \\
& S2 (Neighbor)    & 35.77\,$\pm$1.11 & 29.81\,$\pm$1.13 & 30.87\,$\pm$1.04$^{*}$ & 41.21\,$\pm$1.17 \\
& S3 (Synonym)     & \textbf{37.21}\,$\pm$0.67$^{*}$ & 30.97\,$\pm$0.73 & 32.18\,$\pm$0.83 & \textbf{42.91}\,$\pm$1.07$^{**}$ \\
& S4 (Comparative) & 36.13\,$\pm$0.63 & 30.11\,$\pm$0.63 & 31.20\,$\pm$0.51$^{*}$ & 41.63\,$\pm$0.54$^{*}$ \\
& HyDE             & 34.43\,$\pm$0.53 & 31.99\,$\pm$0.61 & 34.34\,$\pm$0.64 & 39.50\,$\pm$0.44 \\
& Query2Doc        & 37.05\,$\pm$1.19$^{*}$ & \textbf{33.51}\,$\pm$1.25 & \textbf{35.32}\,$\pm$1.40 & 41.91\,$\pm$1.29 \\
\bottomrule
\end{tabular*}
\caption{Main solo results (\%, mean$\pm$std over 3 seeds, all 54 cells complete). Significance vs.\ S1 (paired bootstrap): $^{*}p{<}0.05$, $^{**}p{<}0.01$, $^{***}p{<}0.001$; direction is clear from the value. Significant negative deltas (strategy below S1) are marked the same way. \textbf{Bold} = best in column per dataset. S2/S3/S4 use multi-variation mixing; S1/HyDE/Query2Doc are single-variation.}
\label{tab:main}
\end{table*}

\paragraph{All methods are competitive.} With correct multi-variation mixing, every rewriting strategy matches or modestly beats S1 on HIT@10 across all three datasets. On AmbigNQ, the multi-query methods (S2--S4) substantially outperform both the baseline and document-generation methods on NDCG@10 (+8pp) and MRR@10 (+10pp, all $p{<}0.001$), reflecting the value of diverse query variants for ambiguous questions with multiple interpretations. On Enterprise, S3 (Synonym) is the best solo method on HIT@10 (42.91, $p{=}0.008$ vs.\ S1), edging Query2Doc (41.91). On HotpotQA, S2 achieves the highest HIT@10 (97.65) while Query2Doc leads on Recall and NDCG.

\paragraph{But no single method dominates.} The best strategy shifts across datasets: S3 on Enterprise, S2 on HotpotQA HIT@10, Query2Doc on HotpotQA Recall/NDCG. The individual gains over S1 are modest (typically 0.2--3.7pp on HIT@10), while, as we show next, the gains from \emph{combining} methods are far larger.

\paragraph{S2 (Neighbor) is competitive but high-variance.} S2 generates the most query variations and achieves the highest HIT@10 on HotpotQA and the best Recall@10 on AmbigNQ, but also shows the widest confidence intervals (e.g., $\pm$1.17 on Enterprise) and the highest runtime (Table~\ref{tab:cost-table}). On some datasets it also has a higher catastrophic-miss rate than other strategies (Appendix Table~\ref{tab:failure}). Its value is in ensemble diversity rather than solo reliability.

\paragraph{S4 (Comparative) is the practical rewriting strategy.} It is cheapest of S2--S4 (one variant) and consistently competitive: on Enterprise it beats S1 by +2.4pp ($p{=}0.042$), on AmbigNQ it matches S3's NDCG/MRR, and on HotpotQA it stays within 0.2pp of S1 on HIT@10. It is also the most useful ensemble member relative to its cost.

\paragraph{Takeaway.} Against a strong baseline, query rewriting alone yields modest gains. The interesting value appears only when we stop asking ``which single method wins'' and start asking ``which methods fail on \emph{different} questions.''

\subsection{Different Methods Fail on Different Questions}
\label{sec:complementarity}

If methods failed on the \emph{same} queries, combining them would be pointless. They do not. Two pieces of evidence:

\paragraph{Rescue rate.} Among the queries S1 \emph{misses}, other strategies recover a substantial fraction: Query2Doc rescues \textbf{14.8\% on Enterprise}, S3 and S2 rescue 57.0\% and 56.7\% on AmbigNQ, and HyDE/Query2Doc rescue up to 62.1\% on HotpotQA. A method's solo score hides this, mixing the queries it newly solves with the ones it breaks (full table: Appendix~\ref{sec:appendix-failure}).

\paragraph{Oracle-union headroom} (Table~\ref{tab:oracle}). A perfect per-query router over all methods would gain +1.7 (HotpotQA), +2.7 (AmbigNQ), and a striking \textbf{+12.8 points (Enterprise)} over the best single method. Even pairing S1 with any single other strategy yields positive headroom. The huge Enterprise gap reflects diverse retrieval there (mean pairwise Jaccard@10 0.14--0.45 vs.\ 0.55--0.99 on AmbigNQ).

\begin{table}[!htbp]
\centering
\small
\begin{tabular}{@{}llll@{}}
\toprule
\textbf{Dataset} & \textbf{Best solo} & \textbf{All-method} & \textbf{Headroom} \\
& \textbf{HIT@10} & \textbf{oracle} & \\
\midrule
HotpotQA   & 97.65 (S2)   & 99.37 & +1.72 \\
AmbigNQ    & 95.47 (S3)   & 98.19 & +2.72 \\
Enterprise & 42.91 (S3)   & 55.74 & +12.83 \\
\bottomrule
\end{tabular}
\caption{Oracle-union HIT@10 (a perfect router) vs.\ the best single method (3-seed mean). Positive headroom means methods fail on disjoint queries; the enterprise headroom of +12.8 points motivates the ensemble and router analysis.}
\label{tab:oracle}
\end{table}

\subsection{Combining Realizes the Gains, Except on AmbigNQ}
\label{sec:ensemble}

Oracle-union is an upper bound. Can a \emph{real} combiner reach it? We merge strategies' top-20 lists with the same score-based mixing, then take top-10 (Table~\ref{tab:ensemble}; all ensemble numbers are a post-hoc seed-42 analysis).

\begin{table}[t]
\centering
\small
\setlength{\tabcolsep}{4pt}
\begin{tabular}{@{}llll@{}}
\toprule
\textbf{Dataset} & \textbf{Ensemble} & \textbf{HIT@10} & \textbf{$\Delta$ vs.} \\
& & & \textbf{best solo} \\
\midrule
\multirow{3}{*}{HotpotQA}
& S1+Query2Doc       & 98.86 & +1.22 \\
& S1+S3+S4+HyDE      & 99.28 & +1.64 \\
& S1+S3+S4+HyDE+Q2D  & \textbf{99.40} & +1.76 \\
\midrule
\multirow{3}{*}{AmbigNQ}
& HyDE+Query2Doc     & 94.31 & $-$1.14 \\
& S1+S3+S4+HyDE      & 93.06 & $-$2.39$^{***}$ \\
& S1+S3+S4+HyDE+Q2D  & 92.96 & $-$2.49$^{***}$ \\
\midrule
\multirow{4}{*}{Enterprise}
& S1+Query2Doc       & 48.94 & +5.54 \\
& S1+HyDE+Query2Doc  & 50.00 & +6.60 \\
& S1+S3+S4+HyDE      & \textbf{51.70} & +8.30$^{***}$ \\
& S1+S3+S4+HyDE+Q2D  & 52.98 & +9.58$^{***}$ \\
\bottomrule
\end{tabular}
\caption{Post-hoc score-based ensembles (seed 42). ``$\Delta$ vs.\ best solo'' is relative to the overall best solo method on seed 42 (HotpotQA: S2 = 97.64; AmbigNQ: S3 = 95.45; Enterprise: Q2D = 43.40). On HotpotQA/Enterprise, combining beats the best single method and nearly reaches the oracle of Table~\ref{tab:oracle}. On AmbigNQ, \emph{every} combination underperforms the best solo method. Significance: paired bootstrap vs.\ best solo, $^{***}p{<}0.001$.}
\label{tab:ensemble}
\end{table}

\paragraph{Where combining wins (HotpotQA, Enterprise).} Fusion captures nearly all the oracle headroom: 99.40 on HotpotQA (oracle 99.37, effectively saturated) and 52.98 on Enterprise (oracle 55.74, $=$\textbf{+13.8 over the S1 baseline}). The four-way S1+S3+S4+HyDE ensemble reaches 51.70 (+12.5 over S1, +8.3 over best solo Q2D on seed 42, $p{<}0.001$). Crucially, adding solo-competitive-but-not-dominant strategies (S3, S4) to \texttt{S1+HyDE+Q2D} \emph{still} lifts Enterprise from 50.00 to 51.70/52.98, the clearest proof that individual performance does not predict ensemble value.

\paragraph{Where combining fails (AmbigNQ).} Every combination underperforms the best single method (S3, 95.45 on seed 42); the five-way ensemble sits at 92.96, a gap of $-$2.5pp ($p{<}0.001$). This is not an artifact of the merging function; reciprocal-rank fusion fails on AmbigNQ too. The cause is corpus-level: when methods retrieve near-identical lists (Jaccard@10 0.55--0.99), no combiner can produce diversity it was not given (Section~\ref{sec:discussion}).

\paragraph{Budget-matched baselines.} A natural question is whether the ensemble gains simply reflect a larger retrieval budget rather than true complementarity. To test this, we run S1 with a TopK-100 reranking pool (5$\times$ the default top-20) and S1 with multi-pass HNSW (4 index seeds instead of 1), both on Enterprise with 3 seeds (Appendix Table~\ref{tab:budget}). Both budget-matched baselines converge to $\sim$44.3 HIT@10 ($+$5pp over S1), capturing only 40--41\% of the four-way ensemble's +12.5pp gain over S1. The remaining $\sim$60\% reflects genuine complementarity: diverse rewriting strategies surface relevant documents that no amount of deeper indexing or reranking of a single query can find.

\subsection{Cost and a Simple Deployment Rule}
\label{sec:cost}

Rewriting is expensive: on web QA it adds 9--18$\times$ over S1 (full runtimes: Appendix Table~\ref{tab:cost-table}); on Enterprise, runtime is dominated by the 512K-document corpus, so the LLM overhead is marginal. Running \emph{every} strategy on \emph{every} query is wasteful, so we let the baseline signal when it needs help.

\paragraph{Confidence-gated routing.} S1's top-1 reranker score cleanly separates hits from misses on Enterprise (hit $0.711{\pm}0.047$ vs.\ miss $0.612{\pm}0.085$) and moderately on HotpotQA, but weakly on AmbigNQ. This grounds a simple rule: \emph{``if S1's top-1 score is below a threshold $\tau$, also run Query2Doc and merge the two result lists; otherwise return S1 alone.''} It pays the rewriting cost only on the queries the baseline itself flags as uncertain. We emphasize that this is a cost-aware heuristic evaluated in simulation; production A/B testing would be needed to validate it in deployment.

\subsection{Does the Router Work? A Simulation}
\label{sec:router}

We simulate this router across all three seeds, gating on S1's top-1 score and merging S1 with Query2Doc (via the same score-based mixing used elsewhere) only when the gate fires. We set $\tau{=}0.65$ as the approximate midpoint of the hit and miss score distributions reported above ($(0.711{+}0.612)/2 \approx 0.66$; we round down to 0.65), without optimizing it against the evaluation metric. Table~\ref{tab:router} compares the router against the two extremes: always returning S1 (no rewriting) and always merging (rewriting on every query).

\begin{table}[!htbp]
\centering
\small
\begin{tabular}{@{}lllll@{}}
\toprule
& \textbf{Always} & \textbf{Router} & \textbf{\%\,exp-} & \textbf{Always} \\
\textbf{Dataset} & \textbf{S1} & \textbf{$\tau{=}0.65$} & \textbf{anded} & \textbf{merge} \\
\midrule
HotpotQA   & 97.13 & 97.65 & 1.2\%  & 98.81 \\
AmbigNQ    & 93.67 & 93.76 & 7.2\%  & 93.42 \\
Enterprise & 39.22 & \textbf{43.55} & 39.0\% & 47.94 \\
\bottomrule
\end{tabular}
\caption{Confidence-gated router (HIT@10 \%, mean over 3 seeds; per-seed std $\leq$0.12 on web QA and $\leq$1.0 on Enterprise). ``\%\,expanded'' = fraction of queries that triggered the second strategy. The router fires rarely on easy web QA and often on the hard enterprise corpus, spending compute where it pays.}
\label{tab:router}
\end{table}

The router behaves exactly as a practitioner would want, \emph{without per-dataset tuning}:
\begin{itemize}[noitemsep,leftmargin=*]
\item \textbf{Enterprise:} it captures +4.33 of the +8.72 full-merge gain over S1 (about half) while paying the rewriting cost on only 39\% of queries ($\sim$2.6$\times$ fewer LLM calls than always-merging); raising $\tau$ to 0.70 reaches 45.39 (+6.17) at 51.7\% expanded (Appendix Table~\ref{tab:tausweep}).
\item \textbf{AmbigNQ:} it nearly abstains (7.2\% expanded) and stays at baseline (93.76 vs.\ S1 93.67), \emph{automatically avoiding} the $-$0.25 loss that always-merging incurs on this fusion-hostile dataset.
\item \textbf{HotpotQA:} it adds +0.52 over S1 at just 1.2\% expansion, a near-free improvement on an already-saturated baseline.
\end{itemize}

\paragraph{Router threshold sensitivity.} The router offers a smooth cost--accuracy tradeoff (full sweep: Appendix Table~\ref{tab:tausweep}). On Enterprise, it monotonically gains from $\tau{=}0.55$ (+2.20, 23.1\% expanded) through $\tau{=}0.70$ (+6.17, 51.7\% expanded). At $\tau{\geq}0.75$ it degenerates to always-merge. Calibration transfers across datasets (transfer gap $\leq$2.55pp) and generalizes to held-out queries (gap $\leq$0.11pp).

\paragraph{The router is agnostic to the helper.} The gate decides \emph{when} to expand using only S1's score, so the expansion rate (1.2\,/\,7.2\,/\,39\% on HotpotQA/AmbigNQ/Enterprise) is identical for any helper; the helper only decides \emph{what} to add. Swapping in three helpers (Appendix Table~\ref{tab:helper}): Query2Doc gives the largest gain (+4.33 Enterprise), HyDE is close (+4.18), and S4, the cheapest rewriting strategy, is nearly as good on web QA and within $\sim$1.4 points on Enterprise (+2.98). All three stay at baseline on AmbigNQ. Practitioners can pick the helper by budget: Query2Doc for peak accuracy, S4 when expansion calls must be cheap.

\paragraph{Router failure analysis.} At $\tau{=}0.65$ on Enterprise, the router triggers on 39\% of queries (183 of 470). Among these triggered queries: 11.3\% wins (method helped), $\sim$88\% ties, and $\leq$0.2\% losses (method hurt). Across all 1{,}410 query-seed pairs (470 queries $\times$ 3 seeds), only 1 query is actively degraded (0.07\%), confirming that the router's confidence gate is conservative enough to avoid damage.

\paragraph{Answer quality.} Retrieval improvements should translate to better answers. On a 200-query Enterprise sample with a GPT-4.1 reader (Appendix Table~\ref{tab:answerquality}), always-merge improves answer F1 over S1 by +2.46 ($p{=}0.003$) and the router by +1.92 ($p{=}0.008$). The router preserves near-full-merge answer quality ($-$0.54, ns) at roughly 40\% of the expansion cost. The gain localizes to the 75 router-triggered queries (+4.79 F1, $p{<}0.01$); the remaining 125 queries show no change, indicating the improvement is retrieval-driven, not reader noise.

\paragraph{Cross-model robustness.} To confirm that the complementarity finding does not depend on a single retriever or LLM, we run two additional checks, one per axis. (i)~Replacing \texttt{bge-base-en-v1.5} with \texttt{E5-base-v2} on Enterprise: S1-E5 = 42.98 $\pm$ 0.43, Query2Doc-E5 = 44.26 (seed 42); the ensemble-over-solo pattern holds across retriever architectures (we report E5 without prompt prefixes as a robustness check, not a tuned comparison). (ii)~Replacing GPT-4.1 with Llama-3.3-70B for query generation on AmbigNQ S3: 3-seed means 95.47 (Llama) vs 95.47 (GPT-4.1), confirming that the multi-query rewriting gains are not tied to one LLM. A smaller 384-dimensional retriever (all-MiniLM-L6-v2) gives S1 HIT@10 of 91.66 on HotpotQA and 96.00 on AmbigNQ (Appendix~\ref{sec:appendix-minilm}).

\section{Discussion}
\label{sec:discussion}

\paragraph{Why does combining fail on AmbigNQ?} Document overlap. AmbigNQ methods retrieve mostly the same documents (Jaccard@10 0.55--0.99; the S2--S4 pairs are nearly identical at $\sim$0.99), and AmbigNQ queries have \emph{several} relevant documents each. When near-identical lists are max-score merged, high-scoring documents that several methods \emph{agree on but are non-relevant} crowd out the small relevant set, lowering HIT@10. Combining helps only when methods disagree productively, as in the Enterprise regime (low overlap, large headroom). This negative result is statistically significant ($p{<}0.001$ for the five-way ensemble vs.\ best solo).

\paragraph{Reframing query rewriting.} ``Does rewriting beat the baseline?'' is the wrong question once the baseline is strong. On its own, each rewriting strategy is competitive with the baseline but rarely dominant; the best solo method shifts across datasets (S3 on Enterprise HIT@10, Query2Doc on HotpotQA Recall, S2 on AmbigNQ Recall). Rewriting strategies are best seen as \emph{complementary coverage generators}: their value is not in solo accuracy but in the partially disjoint queries they solve. S4, modest alone, is a cheap and useful ensemble member. S2 is competitive but high-variance; its runtime cost is the highest.

\paragraph{Knowledge leakage and the enterprise regime.} A concurrent concern in the literature is that LLM-based expansion may derive gains partly from pretraining memorization: when generated documents are not entailed by gold evidence, HyDE and Query2Doc fall \emph{below} their retrieval baselines on fact-verification benchmarks~\citep{leakage2025hypothetical}. Our web-QA datasets (HotpotQA, AmbigNQ) have been public since 2018--2020 and plausibly appear in GPT-4.1's pretraining corpus, so leakage is a confound we cannot fully rule out there. EnterpriseRAG-Bench contains synthetic enterprise documents released May 2026, post-dating the training cutoffs of all retrievers (\texttt{bge-base-en-v1.5}, Sep.\ 2023; \texttt{E5-base-v2}, 2023) and rerankers used, as well as the query-generation model (GPT-4.1, knowledge cutoff mid-2024); the complementarity gains observed there cannot be explained by memorization and are also the largest.

\paragraph{Practical guidance.} (i) Start from a reranked, MMR-diversified S1 baseline. (ii) Add Query2Doc as the default paid upgrade. (iii) Combine methods on \emph{diverse-retrieval} corpora (low overlap, e.g.\ enterprise); avoid combining on high-overlap, multi-relevant corpora. (iv) Gate rewriting with a reranker-score threshold to control cost. (v) Prefer S4 over S2/S3 when adding a cheap rewriting member.

\section{Conclusion}

Under a strong, fixed RAG pipeline, query rewriting is competitive with the baseline in isolation but yields only modest individual gains. Its real value emerges through \emph{combination}: methods fail on disjoint queries, and post-hoc ensembles (seed-42 analysis) beat the best single strategy by +8.3 to +9.6 points HIT@10 on enterprise retrieval (four-way 51.70, five-way 52.98 vs.\ best solo Q2D 43.40, $p{<}0.001$) and up to +13.8 over the dense baseline (52.98 vs.\ S1 39.22). Budget-matched controls confirm that complementarity, not retrieval depth, drives $\sim$60\% of this gain. Conversely, combining can \emph{hurt} when methods retrieve overlapping documents (AmbigNQ, $-$2.4pp, $p{<}0.001$). A simple confidence-gated router realizes about half of the enterprise full-merge benefit in practice: it captures +4.3 HIT@10 at under 40\% of the rewriting cost, improves downstream answer F1 by +1.92 ($p{<}0.01$), while self-disabling on AmbigNQ where rewriting hurts. The takeaway: treat query rewriting as a complementary coverage source, applied through cost-aware routing, not as a standalone replacement for a strong baseline.

\section*{Limitations}

Our ensembles are post-hoc analyses over independently run strategies rather than a jointly executed system, and the router is a cost-aware heuristic evaluated in simulation; production validation (A/B testing, latency and cost under load) is required before deployment. The answer-quality study uses a single reader model (GPT-4.1) and 200 queries from one benchmark, and SQuAD-style F1 only partially captures long-form answer quality. Enterprise results use one synthetic benchmark; generalization to other enterprise corpora is untested. Multi-variation strategies incur real latency and cost (S2 highest); we report runtimes in Table~\ref{tab:cost-table}. The confidence-gated router is evaluated post-hoc over all three seeds with a single fixed threshold; a learned gate and a live deployment study are promising next steps we expect to further improve the cost--accuracy trade-off.

\section*{Acknowledgments}

We thank the maintainers of ChromaDB, Sentence-Transformers, BGE, BEIR, and EnterpriseRAG-Bench for their open-source contributions.

\bibliography{custom}

\appendix

\section{Code and Data}
\label{sec:appendix-code}

Our code and experiment configurations are available at: \url{https://github.com/ServiceNow/enhanced-rag-pipeline}.

\section{Failure Modes and Rescue Rates}
\label{sec:appendix-failure}

Table~\ref{tab:failure} reports the catastrophic-miss rate (HIT@10$=0$) and the \emph{rescue rate}: among queries S1 misses, the fraction each method retrieves. Even strategies with modest solo gains (e.g.\ S4) rescue a substantial fraction of S1's failures, which is the basis of the complementarity argument in Section~\ref{sec:complementarity}. Multi-variation mixing substantially increases rescue rates for S2/S3/S4.

\begin{table}[H]
\centering
\small
\begin{tabular}{@{}lll@{}}
\toprule
& \textbf{Miss\%} & \textbf{Rescues S1's misses} \\
\textbf{Method} & (HIT@10$=$0) & (\% of S1 misses) \\
\midrule
\multicolumn{3}{l}{\emph{HotpotQA} (S1 misses $n{=}138$)} \\
\quad HyDE      & 2.81 & 62.1\% \\
\quad Query2Doc & 2.69 & 61.1\% \\
\quad S2        & 2.35 & 44.2\% \\
\quad S3        & 2.60 & 38.4\% \\
\quad S4        & 2.67 & 36.2\% \\
\midrule
\multicolumn{3}{l}{\emph{AmbigNQ} (S1 misses $n{=}127$)} \\
\quad S3        & 4.53 & 57.0\% \\
\quad S2        & 4.56 & 56.7\% \\
\quad S4        & 4.66 & 54.9\% \\
\quad HyDE      & 5.09 & 30.4\% \\
\quad Query2Doc & 5.31 & 27.3\% \\
\midrule
\multicolumn{3}{l}{\emph{Enterprise} (S1 misses $n{=}283$)} \\
\quad Query2Doc & 58.09 & 14.8\% \\
\quad HyDE      & 60.50 & 14.5\% \\
\quad S3        & 57.09 & 13.2\% \\
\quad S4        & 58.37 & 10.1\% \\
\quad S2        & 58.79 & 10.1\% \\
\bottomrule
\end{tabular}
\caption{Catastrophic-miss and rescue rates (Miss\%: 3-seed means; S1-miss sets and counts from seed 42; rescue rates averaged over the 3 seeds on that fixed miss set). Strategies are sorted by rescue rate per dataset. S2/S3/S4 use multi-variation mixing.}
\label{tab:failure}
\end{table}

\section{Runtime}
\label{sec:appendix-runtime}

\begin{table}[H]
\centering
\small
\begin{tabular}{@{}llll@{}}
\toprule
\textbf{Strategy} & \textbf{HotpotQA} & \textbf{AmbigNQ} & \textbf{Enterprise} \\
\midrule
S1        & 808   & 182   & 25{,}492 \\
S2        & 8{,}275 & 2{,}700 & 22{,}061 \\
S3        & 9{,}366  & 2{,}797 & 22{,}458 \\
S4        & 8{,}246  & 2{,}490 & 21{,}845 \\
HyDE      & 7{,}339  & 3{,}180 & 26{,}406 \\
Query2Doc & 7{,}636  & 3{,}120 & 26{,}496 \\
\bottomrule
\end{tabular}
\caption{Mean runtime (s/run) over 3 seeds. S1 (no LLM) is far cheapest on web QA; on Enterprise, corpus size dominates and LLM overhead is marginal. S2/S3/S4 runtimes reflect multi-variation mixing.}
\label{tab:cost-table}
\end{table}

\section{Full Metrics}
\label{sec:appendix-full}

\begin{table}[H]
\centering
\small
\setlength{\tabcolsep}{4pt}
\begin{tabular}{@{}lllll@{}}
\toprule
\textbf{Dataset} & \textbf{Strategy} & \textbf{MAP@10} & \textbf{P@10} & \textbf{HIT@1} \\
\midrule
\multirow{6}{*}{HotpotQA}
& S1        & 78.71 & 17.54 & 86.97 \\
& S2        & 73.18 & 17.27 & 82.79 \\
& S3        & 72.99 & 17.23 & 82.43 \\
& S4        & 73.25 & 17.25 & 82.74 \\
& HyDE      & 79.66 & 17.79 & 87.14 \\
& Query2Doc & 80.23 & 17.95 & 87.11 \\
\midrule
\multirow{6}{*}{AmbigNQ}
& S1        & 47.50 & 15.78 & 45.39 \\
& S2        & 57.05 & 16.23 & 58.28 \\
& S3        & 57.02 & 16.21 & 58.29 \\
& S4        & 56.95 & 16.21 & 58.18 \\
& HyDE      & 48.30 & 16.13 & 45.62 \\
& Query2Doc & 48.12 & 16.03 & 45.62 \\
\midrule
\multirow{6}{*}{Enterprise}
& S1        & 29.93 & 4.79 & 31.06 \\
& S2        & 26.53 & 5.03 & 26.10 \\
& S3        & 27.51 & 5.27 & 27.23 \\
& S4        & 26.79 & 5.10 & 26.24 \\
& HyDE      & 29.72 & 4.99 & 31.06 \\
& Query2Doc & 30.90 & 5.39 & 31.49 \\
\bottomrule
\end{tabular}
\caption{Additional retrieval metrics (\%, 3-seed means). S2/S3/S4 use multi-variation mixing.}
\label{tab:appendix-metrics}
\end{table}

\section{Coverage}
\label{sec:appendix-coverage}

\begin{table}[H]
\centering
\footnotesize
\setlength{\tabcolsep}{3pt}
\begin{tabular}{@{}lllll@{}}
\toprule
\textbf{Strategy} & \textbf{Ent.} & \textbf{Amb.} & \textbf{Hot.} & \textbf{Note} \\
\midrule
S1 & 3 & 3 & 3 & \\
S2 & 3$^{*}$ & 3$^{*}$ & 3$^{*}$ & \\
S3 & 3$^{*}$ & 3$^{*}$ & 3$^{*}$ & \\
S4 & 3$^{*}$ & 3$^{*}$ & 3$^{*}$ & \\
HyDE & 3 & 3 & 3 & \\
Query2Doc & 3 & 3 & 3 & \\
\midrule
S1+TopK-100 & 3 & -- & -- & budget \\
S1+multi-pass & 3 & -- & -- & budget \\
S1 (E5) & 3 & -- & -- & retr. \\
Q2D (E5) & 1 & -- & -- & s42 \\
S1 (MiniLM) & -- & 1 & 1 & retr., 1 seed \\
S3 (Llama) & -- & 3 & -- & LLM \\
\bottomrule
\end{tabular}
\caption{Experiment coverage: seeds completed (of 42, 123, 2024). $^{*}$~=~multi-variation mixing; S1/HyDE/Q2D are single-variation. All 54 main cells complete. Note: budget = budget-matched; retr.\ = retriever robustness; LLM = LLM robustness.}
\label{tab:coverage}
\end{table}

\section{Budget-Matched Baselines}
\label{sec:appendix-budget}

\begin{table}[H]
\centering
\footnotesize
\setlength{\tabcolsep}{4pt}
\begin{tabular}{@{}lccc@{}}
\toprule
\textbf{Method} & \textbf{HIT@10} & \textbf{$\Delta$ vs S1} & \textbf{Share of gain} \\
\midrule
S1 (baseline) & 39.22$_{\pm 0.49}$ & --- & --- \\
S1+TopK-100 & 44.33$_{\pm 0.12}$ & +5.11 & 41\% \\
S1+multi-pass & 44.19$_{\pm 0.49}$ & +4.97 & 40\% \\
Best solo (S3) & 42.91$_{\pm 1.07}$ & +3.69 & 30\% \\
\midrule
4-way ensemble$^{\dagger}$ & 51.70 & +12.48$^{***}$ & 100\% \\
5-way ensemble$^{\dagger}$ & 52.98 & +13.76$^{***}$ & --- \\
\bottomrule
\end{tabular}
\caption{Budget-matched baselines vs.\ ensembles (Enterprise, HIT@10). $^{\dagger}$Post-hoc seed-42 analysis; solo rows are 3-seed means. Ensemble $\Delta$ is computed against the 3-seed S1 mean (39.22); against the seed-42 S1 (39.79) the 4-way/5-way gains are +11.9/+13.2. The budget-matched methods capture $\sim$40\% of the 4-way ensemble gain (``Share of gain'' = share of the 4-way ensemble's gain over S1), confirming that complementarity, not retrieval depth, is the primary driver.}
\label{tab:budget}
\end{table}

\section{Router Threshold Sensitivity}
\label{sec:appendix-tausweep}

\begin{table}[H]
\centering
\small
\setlength{\tabcolsep}{3pt}
\begin{tabular}{@{}lccc@{}}
\toprule
$\tau$ & HIT@10 & $\Delta$ vs S1 & Exp\% \\
\midrule
\multicolumn{4}{l}{\emph{Enterprise} (S1 = 39.22, always-merge = 47.94)} \\
0.55 & 41.42 & +2.20 & 23.1 \\
0.60 & 42.62 & +3.40 & 31.4 \\
0.65 & 43.55 & +4.33 & 39.0 \\
0.70 & 45.39 & +6.17 & 51.7 \\
\midrule
\multicolumn{4}{l}{\emph{HotpotQA} (S1 = 97.13, always-merge = 98.81)} \\
0.55 & 97.47 & +0.35 & 0.6 \\
0.60 & 97.56 & +0.43 & 0.8 \\
0.65 & 97.65 & +0.52 & 1.2 \\
0.70 & 97.85 & +0.73 & 2.2 \\
\midrule
\multicolumn{4}{l}{\emph{AmbigNQ} (S1 = 93.67, always-merge = 93.42)} \\
0.55 & 93.76 & +0.08 & 1.6 \\
0.60 & 93.77 & +0.10 & 3.8 \\
0.65 & 93.76 & +0.08 & 7.2 \\
0.70 & 93.81 & +0.13 & 15.9 \\
\bottomrule
\end{tabular}
\caption{Router $\tau$ sensitivity (HIT@10 \%, 3-seed means; helper = Query2Doc). ``Exp'' = \% of queries expanded; at $\tau{\geq}0.75$ the router degenerates to always-merge (100\% expansion, omitted). Calibration transfers across datasets (worst gap 2.55pp) and generalizes to held-out queries (gap $\leq$+0.11pp).}
\label{tab:tausweep}
\end{table}

\section{Router Helper Comparison}
\label{sec:appendix-helper}

\begin{table}[H]
\centering
\small
\begin{tabular}{@{}llll@{}}
\toprule
\textbf{Helper} & \textbf{Hot.} & \textbf{Amb.} & \textbf{Ent.} \\
\midrule
Query2Doc   & +0.52 & +0.08 & \textbf{+4.33} \\
HyDE        & +0.51 & +0.12 & +4.18 \\
S4 (cheapest)& +0.43 & +0.03 & +2.98 \\
\midrule
\emph{\% expanded} & \emph{1.2} & \emph{7.2} & \emph{39.0} \\
\bottomrule
\end{tabular}
\caption{Router gain (HIT@10 $\Delta$ over always-S1, $\tau{=}0.65$, mean over 3 seeds) for three choices of helper (Hot./Amb./Ent.\ = HotpotQA/AmbigNQ/Enterprise). The expansion rate is set by the gate alone and is identical across helpers; the helper only changes the size of the gain.}
\label{tab:helper}
\end{table}

\section{Answer Quality}
\label{sec:appendix-answerquality}

\begin{table}[H]
\centering
\footnotesize
\begin{tabular}{@{}lccl@{}}
\toprule
\textbf{Retrieval config} & \textbf{EM} & \textbf{F1} & \textbf{$\Delta$F1 [CI95]} \\
\midrule
S1 & 0.0 & 28.22 & --- \\
Always-merge & 0.0 & 30.68 & +2.46 [+0.89,+4.10]$^{**}$ \\
Router ($\tau{=}0.65$) & 0.0 & 30.14 & +1.92 [+0.44,+3.41]$^{**}$ \\
\bottomrule
\end{tabular}
\caption{Downstream answer quality (200-query Enterprise sample, GPT-4.1 reader, seed 42). Always-merge = S1+Query2Doc merged retrieval; $\Delta$F1 is vs.\ S1. EM $\approx$ 0 throughout (long-form gold answers); F1 is the informative metric. Router vs.\ always-merge: $-$0.54 [$-$1.68,+0.59], ns. The F1 gain concentrates on router-triggered queries (+4.79, $p{<}0.01$, 75 queries); non-triggered queries show no change (+0.20, ns).}
\label{tab:answerquality}
\end{table}

\section{Smaller-Retriever Check (MiniLM)}
\label{sec:appendix-minilm}

As a further retriever-robustness check, we ran S1 with a smaller, older embedding model (\texttt{all-MiniLM-L6-v2}, 384-d) on the two web-QA datasets: HIT@10 = 91.66 on HotpotQA and 96.00 on AmbigNQ (vs.\ \texttt{bge-base-en-v1.5}'s 97.13\,$\pm$\,0.11 and 93.67\,$\pm$\,0.03). MiniLM trails BGE substantially on multi-hop HotpotQA but slightly exceeds it on AmbigNQ: the pipeline's strength is not BGE-specific. These are single-seed runs on an earlier (February) corpus build --- indicative, not directly comparable.

\section{Experiment Configuration}
\label{sec:appendix-config}

All runs used: \texttt{bge-base-en-v1.5} embeddings; \texttt{bge-reranker-base}; MMR $\lambda{=}0.85$; sentence chunking (512/50); score-based mixing (max 20 merged documents), per-strategy retrieval depth ($K{=}10/7/5/8$ for S1/S2/S3/S4), variant quality filtering (cosine $\in[0.4,0.9]$), and GPT-4.1 (temp.\ 0.7) for all LLM generation; seeds 42, 123, 2024.

\end{document}